# Analyzing Information-Seeking Behaviors in a Hakka AI Chatbot: A Cognitive-Pragmatic Study

Chu-Hsuan Lee
Department of Cultural Creativity and Digital Marketing, National United University
hsuan6389@nuu.edu.tw

Chen-Chi Chang
Department of Cultural Creativity and Digital Marketing, National United University
kiwi@gm.nuu.edu.tw

Hung-Shin Lee
United Link Co., Ltd.
hungshinlee@gmail.com

Yun-Hsiang Hsu
Department of Cultural Creativity and Digital Marketing, National United University
yuripeyamashita@gmail.com

Ching-Yuan Chen
Department of Cultural Creativity and Digital Marketing, National United University
minicr1234@gmail.com

## Abstract

*With many endangered languages at risk of disappearing, efforts to preserve them now rely more than ever on using technology alongside culturally informed teaching strategies. This study examines user behaviors in TALKA, a generative AI-powered chatbot designed for Hakka language engagement, by employing a dual-layered analytical framework grounded in Bloom's Taxonomy of cognitive processes and dialogue act categorization. We analyzed 7,077 user utterances, each carefully annotated according to six cognitive levels and eleven dialogue act types. These included a variety of functions, such as asking for information, requesting translations, making cultural inquiries, and using language creatively. Pragmatic classifications further highlight how different types of dialogue acts—such as feedback, control commands, and social greetings—align with specific cognitive intentions. The results suggest that generative AI chatbots can support language learning in meaningful ways—especially when they are designed with an understanding of how users think and communicate. They may also help learners express themselves more confidently and connect with their cultural identity. The TALKA case provides empirical insights into how AI-mediated dialogue facilitates cognitive development in low-resource language learners, as well as pragmatic negotiation and socio-cultural affiliation. By focusing on AI-assisted language learning, this study offers new insights into how technology can support language preservation and educational practice.*

## 1. Introduction

The rapid development of generative artificial intelligence (AI) has introduced new opportunities for interactive language learning and cultural engagement. Chatbots, in particular, have gained attention in education for enabling timely interaction, personalized learning, and social presence (Huang et al., 2022). Despite these benefits, most applications have centered on high-resource languages, while low-resource and endangered languages remain underrepresented.

The Hakka language, once a vibrant cultural marker in Taiwan, now faces declining intergenerational transmission and weakening community use. According to the Hakka Affairs Council (2022), only 38.3% of the population are fluent speakers, reflecting a decline compared with 2016. Research has further documented the erosion of Hakka usage in multilingual households, where dominant languages often prevail (Lai, 2017). Revitalization thus requires both linguistic preservation and the reinforcement of cultural identity through daily use.

Recent scholarship emphasizes that AI-driven tools can support these goals but raises concerns about technological limitations, ethical issues, and learner agency (Godwin-Jones, 2024). Existing literature has predominantly focused on AI language learning tools' technical performance and pedagogical effectiveness. At the same time, analysis of how users interact with these systems remains insufficient regarding their questioning patterns, pragmatic behaviors, and cognitive processes (Kuhail et al., 2023). Yang and Kyun (2022) also pointed out that future research should place greater emphasis on the "interaction details" and "dialogue quality" in AI-supported language learning journeys. In particular, understanding the cognitive levels and contextual operational skills reflected in questioning behavior is crucial for comprehending the role of generative AI in the language acquisition process.

Furthermore, Huang et al. (2022) argued that a detailed analysis of learner engagement is necessary for AI tools to be effectively implemented. Biases in large language models (LLMs) highlight the importance of understanding how user inputs shape system responses, particularly for non-mainstream languages (Blasi et al., 2022; Benkirane et al., 2024). According to Korzynski et al. (2023), a lack of clear and purposeful prompting can result in answers that are misleading or misaligned with user expectations. Lewis (2025) further argued that AI language tools without cultural sensitivity and contextual responsiveness reduce minority learners' engagement, underscoring the need for stronger design and user support.

This study addresses these gaps by analyzing user interactions with TALKA, a generative AI chatbot for Hakka language engagement, using a dual framework that integrates dialogue act categorization with Bloom's Taxonomy. A large language model was applied for preliminary classification, with expert validation establishing an 11-category pragmatic scheme. This approach examines communicative functions and cognitive depth in user questions, providing insights into AI-mediated dialogue that inform language preservation, educational design, and inclusive technology development.

# 2. Literature Review

## 2.1 Information Seeking Behavior

With rapid technological growth and broader Internet access, interactions with information have become seamless. Information-seeking behavior refers to the process by which individuals actively search for and filter information from various sources to meet specific needs and goals (Parveaz & Khan, 2022), covering a series of actions in identifying information needs, searching and using information, and even in the process of information transmission (Wilson, 1999). Wilson (1997) noted that meeting information needs drives user action. Learners seek sources based on perceived needs and may restart searches depending on outcomes. In language learning with human–computer interaction, information seeking becomes both a means of knowledge acquisition and a reflection of cognitive and pragmatic intent. Whether in learning-oriented search tasks or the context of human-computer interaction, it can be used as an important observation indicator to understand their learning process and pragmatic strategies (Ghosh et al., 2018; Ghosh & Shah, 2017). In language learning, learners tend to adjust the form of questions according to their familiarity with the task and language ability. When faced with unfamiliar tasks, they often use questioning strategies such as clarification, verification, and confirmation to promote the grasp of semantics and knowledge construction (Mackey et al., 2007). Related research also points out that explicit training on these strategies can help improve second language learners' questioning ability and text comprehension depth (Wang & Wang, 2013). This process also reflects the dynamic operation of pragmatic strategies. Especially in the generative AI dialogue environment, prompting is often regarded as a key factor affecting the quality of generative AI responses in different fields (Knoth et al., 2024; Sikha et al., 2023). In addition, users from different cultural and ethnic backgrounds often show the extension of cultural knowledge and the expression of identity in their information-seeking process. They are influenced by language, social networks, and educational concepts, forming a culturally aware information interaction model (Liu, 1995). Information seeking behavior is not only the starting point of language interaction but also an important basis for observing questioning strategies and pragmatic participation.

## 2.2 Cultural and Cognitive Dimensions

To establish the theoretical foundation of this study's analytical framework, the following section will discuss questioning behavior in the language learning process, focusing on its cognitive levels and pragmatic functions, and further examine the research context of relevant classification systems and cultural sensitivity. In the application fields of language learning and cultural transmission, questioning behavior is a key indicator of a learner's thinking level and linguistic operational ability (Suhardiana, 2019). In recent years, AI-based language learning methods have shown higher learning efficacy than traditional teaching models, particularly demonstrating significant advantages in promoting learners' cognitive development and learning motivation (Javaid, 2024). Although generative AI is considered a technological tool with great potential for language learning, its effectiveness in pragmatic functions and cultural context understanding still falls short of meeting the demands of authentic language interaction (Godwin-Jones, 2024). Meanwhile, there is a relative lack of research on the evolution of learners' pragmatic purposes and cognitive levels during interactions. Therefore, this paper intends to use user data from the TALKA Hakka chatbot to analyze the cognitive depth and cultural-pragmatic strategies implicit in users' active questioning behaviors.

The rapid advancement of digital technologies has prompted the adaptation of Bloom’s Taxonomy to digital learning contexts, where it now serves as a central theoretical foundation for assessing knowledge (Amin & Mirza, 2020). To thoroughly investigate user questioning behavior in the context of generative AI, this study adopts a dual analytical framework that combines Bloom’s Taxonomy with pragmatic function categorization, enabling a systematic examination of both cognitive levels and cultural orientations in learners’ linguistic output. Initially introduced by Bloom (1956), the taxonomy organizes the learning process into six hierarchical levels of cognitive complexity: "Remembering" (recalling knowledge and facts), "Understanding" (explaining meaning), "Applying" (using acquired knowledge in practice), "Analyzing" (breaking down concepts and structures), "Evaluating" (making judgments and choices), and "Creating" (integrating and generating new content). This hierarchical framework provides a structured means to evaluate learners' thinking abilities, ranging from factual recall to higher-order reasoning and creativity, and it has been widely used in language as well as education research to evaluate cognitive depth and knowledge construction (Chandio et al., 2016; Krathwohl, 2002). AI-powered systems have further demonstrated the ability to generate and evaluate questions aligned with Bloom's six cognitive levels, thereby improving the reliability of assessments and the efficiency of learning processes (Yaacoub et al., 2025). Automated classification frameworks like AutoBloom have also illustrated how instructional materials can be systematically aligned with Bloom's Taxonomy, yielding actionable insights for curriculum design (Shaik et al., 2023). While these strengths underscore Bloom's systematic and broad applicability as a benchmark for AI-enhanced language learning evaluation, limitations persist: higher-order levels such as "Creating" remain difficult for AI systems to capture fully, and the taxonomy's cognitive focus tends to neglect affective and psychomotor domains (Elim, 2024; Hui, 2025; Krathwohl, 2002). Furthermore, research on inquiry-based learning shows that learners can enhance contextual understanding and strengthen concept construction through continuous questioning and problem refinement. This process contributes to the development of metacognition and deep learning (Lombard & Schneider, 2013). Since questioning behavior simultaneously reflects learners' cognitive engagement and pragmatic intent, integrating Bloom's hierarchical model with dialogue act analysis is theoretically justified. Bloom's model provides a structured perspective for evaluating the depth of cognitive processing, while dialogue act categorization captures the communicative functions and cultural orientations of language use. Together, they constitute a complementary framework that more comprehensively accounts for both the cognitive and pragmatic dimensions of learner interactions (Anderson & Krathwohl, 2001; Chi, 2009).

On the other hand, recent culture-oriented pragmatic research indicates that minority language users often integrate cultural knowledge, social role awareness, and ethnic identity in their questioning processes, forming highly culturally sensitive patterns of language use. Brixey (2025) further points out that for minority language users, interaction with an AI chatbot is not merely a tool for language practice but also a process of cultural practice and identity construction. When the content of questions involves cultural allusions, folk etiquette, or linguistic landscapes, the linguistic act transcends mere practice, becoming a form of cultural participation and identity expression. Hinkel (2013) notes that language learners in second-language communities often need to understand and adapt to cultural-pragmatic norms. Their questioning behavior often reflects an exploration of the target culture's interaction habits, helping to position their roles and identities in a cross-cultural context. Furthermore, Preksha and Kaur (2024) have confirmed that cultural literacy significantly promotes pragmatic sensitivity, making learners' linguistic output more aligned with specific contexts and sociocultural language rules.

The learner's questioning behavior in the context of generative AI reflects their linguistic operations and cognitive levels. It may also mirror their journey of cultural cognition and identity exploration. The dual framework of Bloom's cognitive levels and a cultural-pragmatic orientation provides a theoretical basis for understanding such questioning behaviors, aiding in classifying different levels of cognitive activities and pragmatic functions within the user corpus. This theoretical perspective also lays the foundation for the classification system and semantic criteria required for the subsequent analysis of the actual TALKA interaction data.

## 2.3 Pragmatic and Dialogic Dimensions

As Generative Artificial Intelligence and LLMs have rapidly developed, dialogue systems are no longer limited to simulating linguistic responses but have gradually evolved into learning tools with the potential for pragmatic interaction. On the level of pragmatic guidance, prompting is effectively enhancing the contextual alignment of language output. The "AI-PROMPT framework" proposed by Korzynski et al. (2023). Further indicates that structured prompting not only improves the accuracy of system-generated content but also helps enhance contextual relevance and the professional level of pragmatic performance. Research by Saks and Leijen (2018) also confirms that learning interventions assisted by prompting can effectively encourage learners to apply cognitive and metacognitive strategies during tasks, thereby improving their self-regulation abilities and learning outcomes. Even if it does not directly impact learning results, the self-regulation process it triggers can influence strategy use and interaction quality during the learning process (Gentner & Seufert, 2020), which can be further inferred to have a positive benefit on pragmatic awareness and contextual adaptability. In sum, Prompting can be seen as a strategic tool for modulating pragmatic input, strengthening contextual responses, and guiding the system to produce more deeply interactive output, playing a key role in generative AI.

Regarding system pragmatic quality and interactional authenticity, the pragmatic quality of a dialogue generation system directly affects interaction effectiveness. Through a Discourse Completion Task (DCT) analysis, Lee & Cook (2024) pointed out that while ChatGPT can currently perform basic request and apology patterns, its pragmatic strategies tend to be simplified, and its style lacks a human-like quality, making it challenging to recognize socio-contextual cues. At the same time, in the absence of cultural context, generative AI tends to reproduce the biases of mainstream languages. Although some research indicates that generative AI has the potential for language teaching, its cultural-pragmatic processing capabilities and contextual appropriateness must be carefully evaluated in practical applications (Godwin-Jones, 2024). This is especially true when dealing with minority languages and diverse cultural backgrounds, as its statistical generation logic may struggle to reflect the complexity of linguistic identity and social context fully. In low-resource language applications, if the system fails to consider the pragmatic characteristics and cultural context of an ethnic language, it may affect dialogue quality and learning outcomes. Taking the Choctaw–English bilingual chatbot proposed by Brixey & Traum (2020) as an example, their research showed that incorporating cultural stories and language into dialogues helped to strengthen users' language practice and cultural participation, further highlighting the core value of culturally sensitive design in language learning interactions.

Synthesizing the literature above, it is evident that effective prompting strategies help guide AI systems to generate linguistic output with greater contextual depth and interactional meaning, while pragmatic quality and cultural sensitivity are directly related to users' level of dialogue participation and learning effectiveness. However, existing research has focused chiefly on high-resource languages like English. There remains a lack of systematic empirical investigation into how low-resource language learners exhibit questioning strategies, pragmatic operations, and cultural participation in generative AI conversations. Using the TALKA system as an empirical platform, this study aims to explore the interactional characteristics of user questioning behavior at the intersection of cognitive levels and cultural-pragmatic contexts, hoping to address the gap in the existing literature on low-resource language interaction research.

# 3. Methodology

## 3.1 Research Design

This study employs a mixed-methods design, integrating quantitative text classification with qualitative pragmatic interpretation. A dual-axis matrix combining Dialogue Act Categorization and Bloom's Taxonomy was applied to map user utterances across dialogue functions and levels of cognitive complexity. Language learning is conceptualized as both linguistic transmission and the expression of intent, judgment, and creativity in authentic contexts. Analysis of TALKA interactions examines how dialogue types (e.g., greetings, translation requests, cultural questions, feedback) correspond to cognitive activities ranging from recall to creative generation.

## 3.2 Data Collection

The dataset consists of 7,077 utterances from TALKA, a Hakka AI chatbot available via the LINE platform in Taiwan. Data were collected between January and June 2025 and anonymized in compliance with institutional ethical standards. Each record includes user input, timestamp, language mode (Mandarin or Hakka), and system metadata. Preprocessing removed commands, duplicates, and non-linguistic noise, followed by stratified sampling to ensure representation across pragmatic categories. The corpus preserves both spontaneous and task-oriented interactions, supporting cognitive–pragmatic analysis.

## 3.3 Data Annotation and Categorization

To investigate the cognitive and pragmatic dimensions of user interactions with the TALKA chatbot, this research utilized a two-layer annotation framework that combined Bloom's Revised Taxonomy of Cognitive Processes with a dialogue act categorization system. This dual approach enabled the systematic classification of utterances based on the user's cognitive engagement and communicative intention. This study implemented cognitive tagging through an AI-assisted approach, automatically generating preliminary labels for Bloom's cognitive dimensions (Knowledge, Comprehension, Application, Analysis, Evaluation, Creation). Two domain experts reviewed and verified the AI-generated labels to ensure accuracy and validity. This dual-check procedure enhanced the reliability of the tagging process and secured consistency across the dataset. Illustrative examples of user–chatbot interactions with annotations are provided to facilitate reader comprehension. Table 1 presents selected dialogues, including the original Chinese utterances, their English translations, Hakka chatbot responses, dialogue types, and corresponding Bloom's levels.

Cognitive categorization was based on Bloom's six hierarchical levels: Remembering (K), Understanding (C), Applying (Ap), Analyzing (An), Evaluating (E), and Creating (S/Cr). In addition, two functional categories were introduced to capture utterances that fell outside conventional learning objectives. The Social/Expressive (Soc) category included utterances like greetings, emotional remarks, and other interpersonal signals. In contrast, the Command (Cmd) category captured user inputs that directed the system to perform specific actions or manage interactions. Pragmatic annotation used an inductively refined 11-category schema: (1) Daily Interaction, (2) Information Query, (3) Language Learning & Translation, (4) Creative Task, (5) About Chatbot, (6) Seeking Help, (7) System Command, (8) Feedback, (9) Testing/Gibberish, (10) Inappropriate Content, and (11) Hakka Culture & Customs.

To keep the analysis focused and consistent, each utterance was labeled with a single primary dialogue act category. For cognitive tagging, however, multiple labels were allowed when an utterance reflected more than one type of cognitive activity. These annotations made it possible to build a matrix linking dialogue types with cognitive levels, which served as the foundation for the heatmap and behavioral pattern analysis discussed in the sections that follow.

**Table 1. Annotated examples of user–chatbot interactions**

| User msg_text (Chinese) | User msg_text (English) | Chatbot reply_raw_text (Hakka) | Dialogue Type | Bloom's Level |
| --- | --- | --- | --- | --- |
| 我剛剛上班完很累你可以安慰我嗎 | I just finished work and I'm very tired, can you comfort me? | 無干係，辛苦了！(It's okay, you've worked hard!) | 6.SAH | Ap |
| 你可以幫忙嗎 | Can you help? | 毋毋閒，有閒閒日仔。 (Not always, only when I have free time.) | 5.AC | K |
| 你是男生還是女生? | Are you male or female? | 係細妹。 (I am a girl.) | 5.AC | K |
| 你會唱歌嗎? | Can you sing? | 歌曲名稱：菊花台 (Song title: Chrysanthemum Terrace) | 5.AC | K |
| 你有打過電動嗎? | Have you played video games? | 係个，它係電動个。(Yes, it's a video game.) | 5.AC | K |
| 我要上課了，再見。 | I'm going to class, goodbye. | 正來寮！掰掰。(Alright, bye!) | 1.DI | Soc |
| 很高興認識你欸 | Nice to meet you. | 你好，當歡喜認識你。 (Hello, I'm happy to meet you.) | 1.DI | Soc |

## 3.4 Distribution of Dialogue Types

An initial look at the dataset shows that utterances weren't evenly distributed across the eleven dialogue act categories. As illustrated in Table 2, Daily Interaction and Greetings (DI) was by far the most common, with 2,283 entries—about 32.3% of the total. This high frequency suggests that many users engaged the chatbot in routine social exchanges like greetings, simple openers, or phatic expressions, using it not only as a language learning tool but also as a space for casual, low-pressure interaction.

The second and third most prevalent categories are Information Query (IQ) with 1,071 utterances (15.1%) and Language Learning and Translation (LLT) with 1,053 utterances (14.9%), reflecting the strong utilitarian function of the chatbot as a source of linguistic clarification and factual information. These two categories combined account for nearly one-third of all interactions, underscoring the system's role in supporting early-stage language learners who rely heavily on direct questions and translation support.

Other prominent categories include About Chatbot (AC) at 613 utterances (8.7%), Creative Task (CT) at 457 utterances (6.5%), and Hakka Culture and Customs (HCC) at 378 utterances (5.3%), which together highlight diverse user engagement beyond strictly instructional purposes—spanning metalinguistic reflection, creativity, and cultural inquiry.

Although less common, categories like System Command (SC), Seeking Advice and Help (SAH), and Feedback and Evaluation (FE) still played a meaningful role in the dataset. Together, they reflect users' willingness to explore the system's features and share their thoughts on its performance. By contrast, Testing and Gibberish (TG) and Inappropriate Content (IC) appeared only occasionally—just 2.6% and 2.4% of all utterances, respectively—pointing to a generally high level of user engagement and purposeful interaction.

These frequency trends help set the stage for the cognitive-pragmatic analysis that follows, highlighting the range of user behaviors—not only in terms of linguistic function but also in how users engage with the system on a social and emotional level.

**Table 2. Distribution of User Utterances by Dialogue Act**

| Dialogue Act Category | Frequency |
| --- | --- |
| 1.DI | 2283 |
| 2.IQ | 1071 |
| 3.LLT | 1053 |
| 4.CT | 457 |
| 5.AC | 613 |
| 6.SAH | 319 |
| 7.SC | 340 |
| 8.FE | 212 |
| 9.TG | 184 |
| 10.IC | 167 |
| 11.HCC | 378 |
| Total | 7077 |

### 3.5 Distribution of Cognitive Process Levels

To provide a comprehensive understanding of users' cognitive engagement, this section presents a frequency-based analysis of utterances categorized under Bloom's Taxonomy. As shown in Table 3, Remembering (K) was the most dominant cognitive process, appearing in 3,001 of the 7,077 utterances (42.4%). This reflects users' frequent reliance on the chatbot for vocabulary recall, factual queries, and translation—typical of early-stage language acquisition.

Application (Ap) followed with 800 utterances (11.3%), highlighting users' practical use of learned vocabulary or grammar. Evaluation (E) ranked third with 641 instances (9.1%), often involving user judgment or feedback on the chatbot's output. Creating (S/Cr) accounted for 583 utterances (8.2%), revealing user engagement in open-ended creative tasks like idiom construction or poem generation. Comprehension (C) occurred 389 times (5.5%), while Analysis (An) remained relatively rare (184 instances, 2.6%), likely reflecting the advanced proficiency such tasks require.

This distribution reflects a full spectrum of cognitive engagement, from surface-level recall to deeper interpretive and generative behaviors. Notably, higher-order processes such as evaluation and creation underscore the potential of generative AI tools like TALKA to facilitate meaningful interaction and cultural expression in low-resource language contexts.

**Table 3. User Utterances by Cognitive Level**

| Cognitive Level | Total Count |
|---|---|
| Knowledge (K) | 3001 |
| Comprehension (C) | 389 |
| Application (Ap) | 800 |
| Analysis (An) | 184 |
| Synthesis / Creating (S/Cr) | 583 |
| Evaluation (E) | 641 |

### 3.6 Cognitive–Pragmatic Matrix Framework

The analysis proceeded by aggregating the annotated utterances into a two-dimensional matrix, crossing the eleven dialogue act categories with the eight cognitive process levels derived from the extended Bloom's framework. This matrix served as the analytical foundation for identifying patterns in user behavior, examining the distribution of cognitive levels across pragmatic functions, and exploring the relationship between dialogue intentions and cognitive engagement.

Frequency distributions were calculated for each cell in the matrix, and heatmaps were generated to visualize the intensity and spread of cognitive activity within each dialogue act category. This allowed for the identification of cognitive hotspots—areas where users exhibited higher-order thinking such as evaluation and creation—as well as low-engagement zones dominated by recall or social routines.

To explore the pedagogical and cultural implications of these patterns, qualitative inspection was also conducted on representative utterances from key cells, especially those involving creative tasks, cultural inquiries, and evaluative feedback. These qualitative excerpts were triangulated with frequency data to enrich the interpretation of user intent and depth of engagement.

Furthermore, the co-occurrence of dialogue types and cognitive levels was examined to infer user learning strategies and behavioral tendencies. For instance, utterances within the "Language Learning and Translation" category were predominantly associated with applying and remembering levels, whereas the "Creative Task" and "Hakka Culture and Customs" categories showed elevated frequencies of creating and evaluating.

This integrative analytical process allowed for the interpretation of not only how users interacted with the system, but also what kinds of cognitive behaviors were elicited by specific types of dialogue functions. The findings offer insights into the pedagogical affordances of generative AI in low-resource language contexts, with implications for chatbot design, language learning, and cultural identity reinforcement.

### 3.7 Data Visualization and Cross-tabulation

To enhance the interpretability of the annotated dataset and facilitate cross-dimensional analysis, a grayscale heatmap was generated to visualize the frequency distribution of utterances across the intersection of dialogue act categories and Bloom's cognitive levels. The annotated corpus was used to build a two-dimensional matrix, which provided the basis for the resulting visualization.

The horizontal axis of the heatmap represents the eight cognitive categories, including the six canonical Bloom levels—Analyzing (An), Applying (Ap),

Comprehending (C), Command (Cmd), Evaluating (E), Remembering (K), Creating (S), and Social/Expressive (Soc)—while the vertical axis displays the eleven dialogue types annotated in the corpus. Each cell in the matrix reflects the number of utterances within a given pairwise combination of a user's conversational act and cognitive level.

Color intensity corresponds to the frequency of utterances, ranging from white (zero occurrences) to black (more than 2,000 occurrences), enabling readers to identify major interaction patterns quickly. For example, the darkest blocks (DI × Soc) reflect the dominance of social greetings in daily interactions. In contrast, medium-gray blocks indicate minor but meaningful behaviors such as translation tasks (LLT × Ap) or knowledge retrieval (IQ × K).

The grayscale palette was chosen for this study to ensure printing compatibility for academic publications and to enhance the contrast between high- and low-frequency regions. The heatmap was rendered using Python’s seaborn library with square cell dimensions and frequency annotations embedded within each cell for precise reference.

This visualization method facilitates both macro-level pattern recognition and micro-level inspection, rendering it an indispensable instrument for discerning the cognitive-pragmatic structure of user interactions with chatbot systems.

## 4. Results and Discussion

This section highlights the main insights from analyzing user utterances on both cognitive and pragmatic levels. It explores what these patterns reveal about how learners engage with the Hakka language through interactions with a generative AI chatbot. The analysis is structured around the distribution patterns observed in the dialogue act × Bloom cognitive level heatmap and interpreted through quantitative frequency and qualitative meaning. The heatmap in Figure 1 presents the distribution of user utterances across a two-dimensional matrix formed by eleven dialogue act categories and eight cognitive levels adapted from Bloom’s Taxonomy.

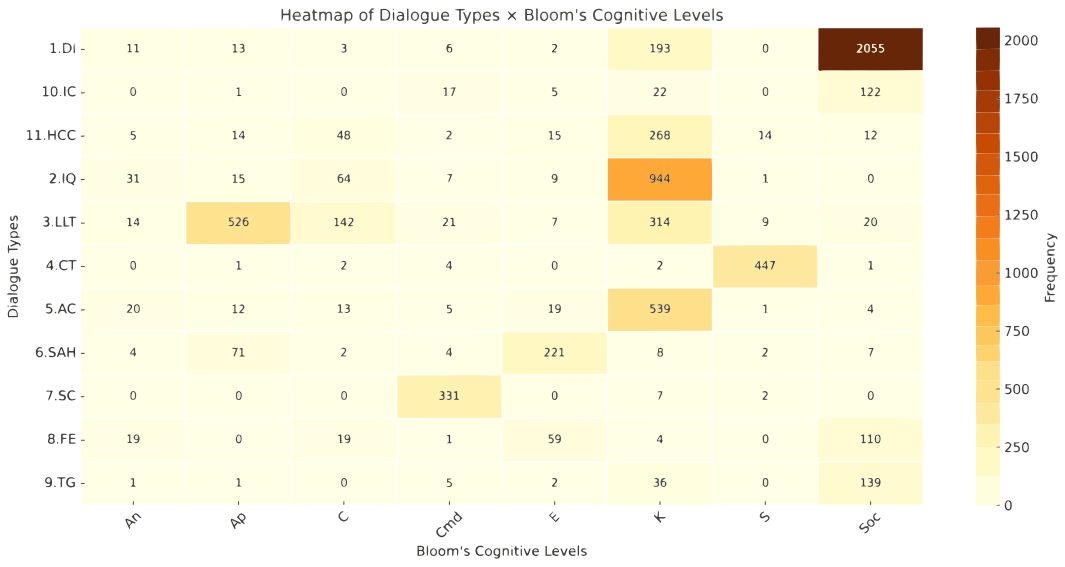

**Figure 1. Heatmap of User Utterance Distribution by Dialogue Type and Bloom’s Levels**

This visualization illustrates the intensity and patterns of user interaction with the Hakka AI chatbot, emphasizing the functional purpose of each utterance and the associated cognitive processes. Notably, high frequencies appear in low-order cognitive categories such as Remembering and Applying, particularly in language learning, information-seeking, and translation-related dialogue. While lower-order tasks are everyday, users also actively engaged in more complex processes like creating and evaluating—particularly when responding to cultural questions or creative prompts. These patterns show the range of learner intentions and behaviors the chatbot can support, revealing how pragmatic use and cognitive depth interact in a low-resource language.

### 4.1 Dominance of Low-Cognitive, High-Social Interactions

As visualized in the matrix heatmap, the highest concentration of user activity occurred in the intersection of “Daily Interaction and Greetings” (DI) and the “Social/Expressive” (Soc) category (n = 2,055). This indicates that users primarily employed the chatbot for interpersonal routines such as greetings, casual phrases, and minimal-content exchanges. While these interactions do not represent deep cognitive processing, they play an important role in lowering the affective barrier to language use, aligning with theories of affective engagement in computer-assisted language learning (CALL). These patterns suggest that social language practices serve as entry points for learners to build familiarity with the system and establish confidence in low-stakes communication.

### 4.2 Active Engagement in Language Application and Retrieval

Substantial user activity was also observed in “Language Learning and Translation” (LLT), particularly at the “Applying” (Ap) and “Remembering” (K) levels, with frequencies of 526 and 314, respectively. Similarly, “Information Query” (IQ) interactions were heavily concentrated at the “Remembering” level (n = 944). These results suggest that learners often tried to recall words, verify translations, or look up factual details—activities typical of early to intermediate stages in second language learning. The connection between these interactions and lower to mid-level cognitive processes highlights how the chatbot is a hands-on resource for building vocabulary and receiving immediate feedback on word usage.

### 4.3 Emergence of Higher-Order Thinking in Creative and Cultural Tasks

The categories “Creative Task” (CT) and “Hakka Culture and Customs” (HCC) exhibited notable activity at higher cognitive levels such as “Creating” (S/Cr), “Evaluating” (E), and “Comprehending” (C). For example, CT × S/Cr yielded 447 utterances, indicating that users actively engaged the chatbot to generate poems, stories, or idiomatic expressions—language use that demands synthesis, patterning, and output beyond simple retrieval. Additionally, HCC × E (n = 15) and HCC × C (n = 48) suggest that users were acquiring cultural knowledge and reflecting upon and critically assessing it. These patterns illustrate the capacity for generative AI chatbots to scaffold deeper linguistic and cultural cognition, particularly when prompts are designed to elicit reflection or cultural storytelling.

### 4.4 System-Oriented Interactions and Feedback

Dialogues categorized under “System Command” (SC) were almost exclusively associated with the “Command” (Cmd) cognitive tag (n = 331), confirming that users were aware of and willing to manipulate the technical affordances of the chatbot system. In addition, “Feedback and Evaluation” (FE) was moderately represented in both the “Evaluation” (E) and “Social” (Soc) categories (n = 59 and 110, respectively), indicating that participants partake in evaluative discussions about the chatbot’s performance, accuracy, or usefulness. These results underscore the emergence of meta-awareness in user–AI interaction, reflecting linguistic behavior, technology critique, and system appraisal.

### 4.5 Cross-Level Complexity and Identity Expression

Some utterances—particularly those in “About Chatbot” (AC) and “Seeking Advice and Help” (SAH)—spanned multiple cognitive levels, including combinations of “Applying,” “Evaluating,” and “Creating.” These hybrid cases reflect exploratory behavior, curiosity, and emotional engagement. In several instances, users expressed identity-related concerns, such as how to speak like an elder, how dialect usage varies across regions, or how to express personal feelings in Hakka. Although not directly coded under an identity model like Dilts’, such expressions signal the chatbot’s emerging role in mediating cultural knowledge and self-positioning within a linguistic heritage.

## 5. Conclusion and Implications

This study examined user interactions with TALKA, a Hakka generative AI chatbot, through a dual framework combining Bloom’s Taxonomy and dialogue act categorization. The analysis produced three main findings. First, user utterances spanned a wide cognitive range, from basic recall and vocabulary use to higher-order evaluation and creative synthesis. Second, pragmatic functions—such as translation, cultural inquiries, and creative prompts—were closely tied to cognitive complexity, showing how dialogue design can shape language processing depth. Third, although less frequent, interactions engaging with identity and culture highlighted the potential of AI systems to foster self-expression and strengthen linguistic heritage in digital contexts.

### 5.1 Theoretical Implications

This research adds to the expanding corpus of research at the intersection of artificial intelligence, language learning, and sociocultural studies. By applying Bloom’s Taxonomy to chatbot-generated user data, the research extends cognitive frameworks into AI-mediated environments, offering a replicable methodology for analyzing educational engagement through naturalistic dialogue. Moreover, including dialogue act categorization broadens the analytical scope, allowing for an integrated view of intent and depth in user expression.

The dual-axis matrix proposed in this study is a novel analytical tool for evaluating user cognition in language technology platforms. It enables the disaggregation of interaction types and supports future research in cognitive modeling, behavior prediction, and adaptive dialogue system design for low-resource languages.

### 5.2 Practical Implications

The results offer concrete insights for designing and deploying AI-based language learning tools. Systems incorporating culturally meaningful content and diverse dialogue types can encourage learners to progress beyond rote memorization and engage in reflection, creative production, and critical inquiry. To better support learning, systems like TALKA could be improved by offering more targeted guidance—helping users move toward deeper thinking—and adapting prompts to match each learner’s progress and needs.

In addition, user-initiated feedback, system control commands, and meta-comments about chatbot behavior point to the importance of transparency, customization, and user agency in conversational agent design. These elements are especially critical when targeting minority language users, for whom trust, relevance and identity affirmation are key to sustained engagement. TALKA incorporates local cultural archives, township documents, and user content into its knowledge database to mitigate possible cultural bias in AI-driven conversations. This RAG-based approach ensures that responses are not solely shaped by generalized patterns in large-scale models but are grounded in region-specific cultural contexts. Moreover, expert validation during the system's design further mitigated the risk of misrepresentation. Given users' strong interest in cultural topics, this design strategy helped enhance the authenticity and cultural sensitivity of TALKA's responses.

### 5.3 Limitations and Future Research

Although annotations provide information about how users use the system, they still have limitations. Since the analysis focused primarily on user input rather than the chatbot's responses, it does not fully capture how system outputs may shape user cognition and conversational trajectories. This selective focus offers a clear view of user-driven behaviors but overlooks the reciprocal dynamics of dialogue. Furthermore, although the categorization frameworks are based on different theories, researchers may need to adjust them when applied to multilingual settings or more complex, multimodal interactions.

Future research could follow this direction by combining conversation analysis of user and system turns, leading to a more comprehensive understanding of cognitive and pragmatic processes in AI-mediated language learning. Longitudinal data could further track learning progress, while evaluating conversation quality from the user and system perspectives, and applying the framework to other low-resource language settings. Further exploration of how generative AI affects users' identity formation and cultural perception would also deepen the sociolinguistic dimension of this study.

## 12. Supplementary Material

A full version of the anonymized TALKA dataset is provided as supplementary material and is accessible at GitHub: https://github.com/yuripeyamashita/talka-user-input